\PassOptionsToPackage{table}{xcolor}
\PassOptionsToPackage{normalem}{ulem}
\documentclass{jingdong}

\usepackage{lmodern}

\renewcommand{\abstractinfont}{\fontsize{9.5}{11.3}\selectfont}
\usepackage[utf8]{inputenc}
\usepackage{url}
\usepackage{tabularx}
\usepackage{amsfonts}
\usepackage{amsmath}
\usepackage{amssymb}
\usepackage{amsthm}
\usepackage{nicefrac}
\usepackage{wrapfig}
\usepackage{enumitem}
\usepackage{pifont}
\usepackage{algorithm}
\usepackage{algpseudocode}
\definecolor{TableHeader}{HTML}{F2F4F5}
\definecolor{TableHighlight}{HTML}{EFF8F1}

\graphicspath{{images/}}

\theoremstyle{plain}

\theoremstyle{definition}

\theoremstyle{remark}

\title{Learning from the Near Future: Temporal Self-Distillation for RLVR}
\author[1,2,*]{Chuanyu Qin}
\author[1,2,*]{Chenxu Yang}
\author[3,*]{Qingyi Si}
\authorbreak
\author[1,2]{Naibin Gu}
\author[1,2]{Dingyu Yao}
\author[1,2]{Zheng Lin}
\author[1,2,\dagger]{Peng Fu}
\author[3]{Nan Duan}
\author[3]{Jiaqi Wang}

\affiliation[1]{Institute of Information Engineering, CAS}
\affiliation[2]{School of Cyber Security, UCAS}
\affiliation[3]{JD.COM}
\contribution[*]{Equal contribution}
\contribution[\dagger]{Corresponding author}

\hypersetup{
  pdftitle={Learning from the Near Future: Temporal Self-Distillation for RLVR},
  pdfauthor={Chuanyu Qin, Chenxu Yang, Qingyi Si, Naibin Gu, Dingyu Yao, Zheng Lin, Peng Fu, Nan Duan, Jiaqi Wang}
}

\abstract{

Reinforcement learning with verifiable rewards (RLVR) is a core post-training recipe for reasoning models, yet pure on-policy learning can be inefficient when useful trajectories are difficult to discover or exploration narrows. 
Existing self-guided approaches largely reuse capability already available to the current or earlier learner. We instead ask whether learning can also make use of capabilities that emerge later in training: \textit{can a model learn from its own future self?}
We introduce \textbf{temporal self-distillation}, in which a policy receives guidance from a stronger later checkpoint of itself. We hypothesize that the most useful temporal teacher need not be the strongest one: a teacher must provide sufficiently new capability while remaining compatible enough for that capability to be readily transferred, motivating a \textbf{near-future} regime.
We study this principle through two complementary mechanisms. \textbf{Near-Future Policy Optimization (NPO)} performs off-policy behavioral transfer using verified future-self trajectories, while \textbf{Near-Future Policy Distillation (NPD)} performs on-policy token-level transfer on learner-generated trajectories. We further introduce \textbf{AutoNPO}, which adaptively determines when temporal guidance is useful and how far to roll back, turning future-self guidance into a repeated self-bootstrap process.
Across eight image-text benchmarks, NPO improves GRPO from 60.25 to 62.84 and AutoNPO reaches 63.15, with consistent gains on text-only and video reasoning. Under NPD, a near-future teacher reaches 63.23 after continued RL versus 61.92 with a far-future teacher, despite lower immediate post-distillation performance. 
Together, these results suggest that effective temporal self-distillation depends not simply on teacher strength, but on a balance between newly acquired capability and learner compatibility.
}

\begin{document}
\maketitle
\vspace{-7mm}

\begingroup
\setlength{\intextsep}{4pt}
\captionsetup{skip=3pt}
\begin{figure}[H]
\centering
\includegraphics[width=0.84\textwidth]{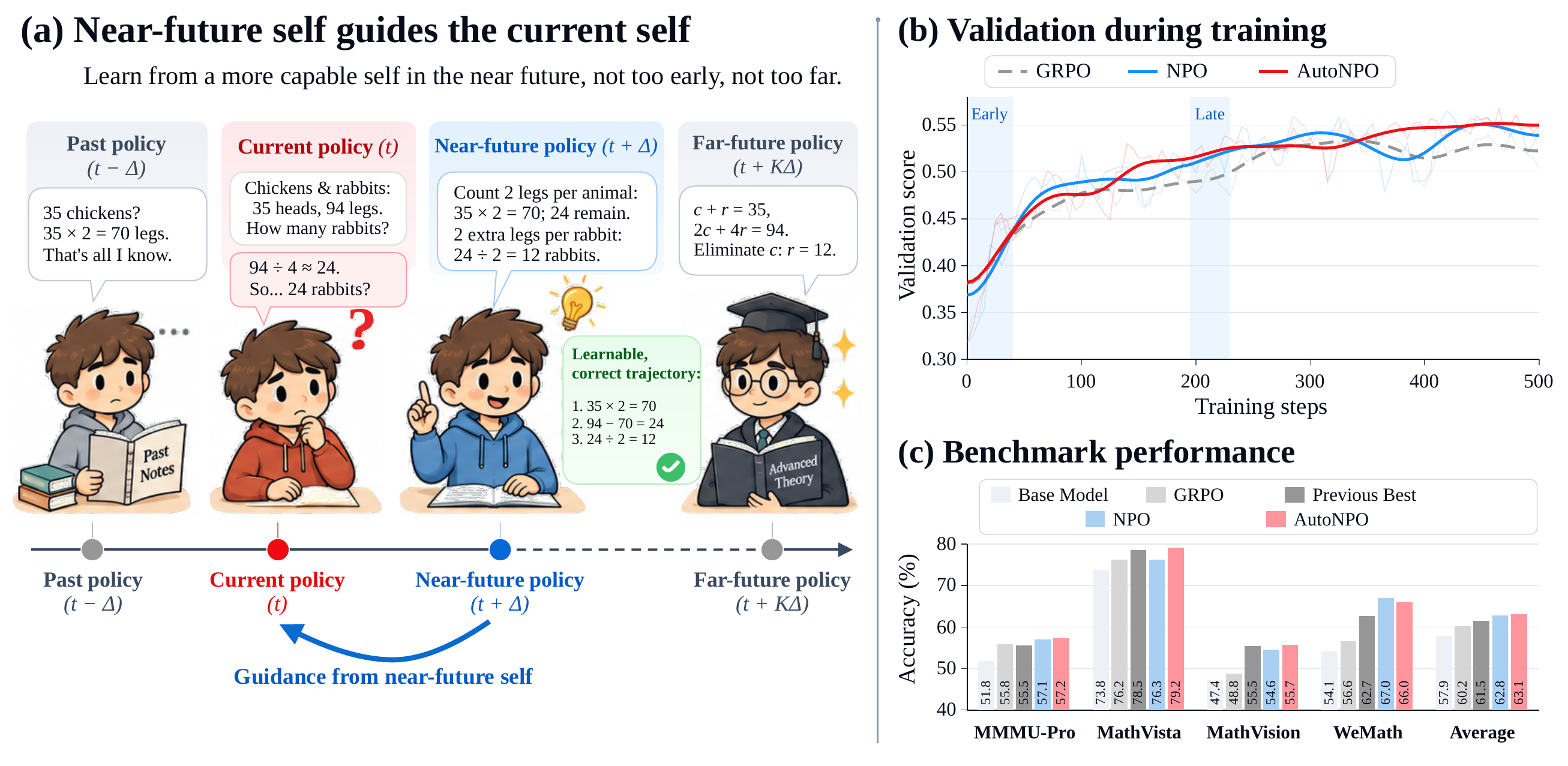}
\caption{
\textbf{(a)}~Guidance from a near-future self.
\textbf{(b)}~Validation curves: thin/thick lines show raw/smoothed scores; shading marks manual interventions.
\textbf{(c)}~Representative image-text results.
}
\label{fig:intro_teaser}
\end{figure}
\endgroup

\clearpage

\section{Introduction}
\label{sec:intro}

Reinforcement learning with verifiable rewards (RLVR) has become a core post-training recipe for reasoning models~\citep{guo2025deepseek,grpo,yu2025dapo}. Yet pure on-policy learning can be inefficient: correct trajectories are scarce early on, while later the rollout distribution may narrow and performance can plateau~\citep{yue2025does,zhang2025rlep}. Existing methods therefore supplement on-policy learning with signals from stronger external models~\citep{yan2025learning,huang2025prefixrft,ma2025learning,fu2025srft} or from experience accumulated during training~\citep{zhan2025exgrpo,zhang2025rlep}. External guidance can provide capabilities beyond the learner's reach but may be difficult to absorb, whereas historical experience remains closer but is limited by what earlier policies had discovered.

A related direction makes learning more self-guided, reusing successful traces~\citep{gulcehre2023reinforced,zelikman2022star} or deriving supervision from the current model conditioned on solutions or feedback~\citep{zhao2026opsd,hubotter2026reinforcement}. Such methods largely look \textit{backward or inward}. Training, however, naturally produces increasingly capable yet closely related policies, with later checkpoints acquiring solutions unavailable to earlier ones. This raises a natural question: \textit{can a model learn from its own future self?} We call this setting \textbf{temporal self-distillation}.

The central problem is then \textit{which future self to learn from}. A teacher that is too close may contribute little new capability, while one that is too far may be difficult to absorb. We characterize this tension through \textbf{signal quality}, which captures useful capability beyond the current policy, and \textbf{guidance cost}, which captures transfer difficulty. As checkpoint distance $\Delta$ grows, capability gain may increase, but so can transfer cost. Our pilot study reveals an intermediate regime where these effects are best balanced, suggesting that a moderately stronger future self may provide more useful guidance than either a very close or a much stronger distant teacher.

We study this principle through two complementary transfer mechanisms. \textbf{Near-Future Policy Optimization (NPO)} performs off-policy behavioral transfer by bringing verified trajectories from a future policy back to an earlier learner while preserving the RL objective. \textbf{Near-Future Policy Distillation (NPD)} builds on on-policy distillation~\citep{agarwal2024policy,lu2025onpolicydistillation}, providing token-level teacher distributions on learner-generated trajectories. Together, they decouple \textit{temporal teacher selection} from the \textit{mechanism of knowledge transfer}, allowing us to test whether the near-future advantage extends beyond off-policy trajectory transfer.

While NPO and NPD assume a chosen temporal teacher, practical training also requires deciding when future guidance is useful and how far to look. We therefore introduce \textbf{AutoNPO}, which automatically determines when to intervene and how far to roll back. AutoNPO turns temporal self-distillation into a repeated self-bootstrap process: later-acquired capability strengthens an earlier learner, which continues training and may produce an even stronger future guide. Thus, improvement itself becomes a reusable learning signal. Although the intervention remains hand-designed, this recursive structure points toward more self-directed improvement.

Empirically, NPO improves the eight-benchmark image-text average from 60.25 to 62.84, while AutoNPO reaches 63.15, with consistent gains on text-only and video reasoning. Under NPD, a near-future teacher reaches 63.23 after continued RL versus 61.92 for a far-future teacher, despite lower immediate post-distillation performance. This cross-mechanism result suggests that the near-future advantage is not merely an artifact of reduced off-policy trajectory shift.


Our contributions are as follows.
\begin{itemize}[leftmargin=1.5em]
    \item We introduce \textbf{temporal self-distillation}, in which a model reuses capabilities acquired by later versions of itself, and frame temporal teacher selection through a trade-off between additional capability and learner compatibility.
    \item We instantiate this principle through \textbf{NPO} for off-policy behavioral transfer and \textbf{NPD} for on-policy distributional transfer, and further introduce \textbf{AutoNPO} to adaptively determine when and how far temporal guidance should be applied.
    \item Across image-text, text-only, and video reasoning, temporal self-distillation consistently improves RLVR, while NPD shows that the near-future advantage persists beyond off-policy trajectory transfer.
\end{itemize}

\begin{figure*}[t]
\centering
\includegraphics[width=0.98\textwidth]{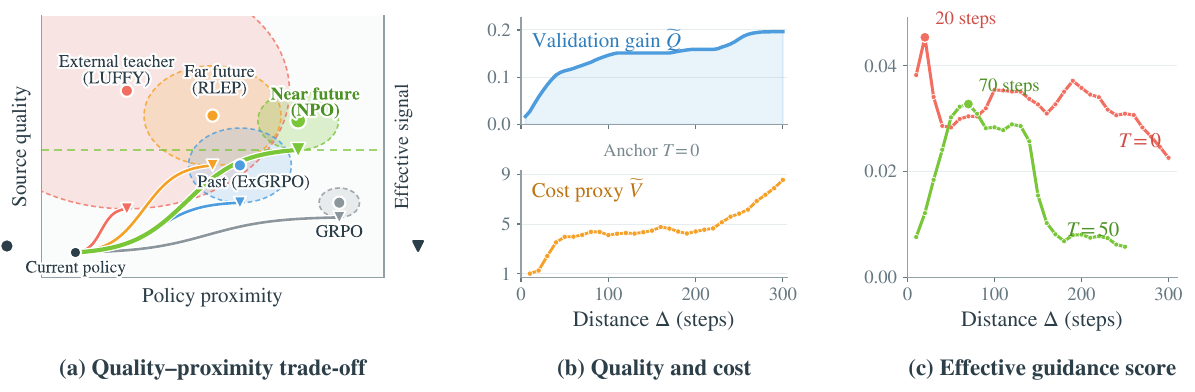}
\caption{\textbf{Why the near future can be a useful guide.}
\textbf{(a)} Conceptual comparison of guidance sources by signal quality and learner compatibility.
\textbf{(b)} Along a GRPO training trajectory, farther checkpoints can provide stronger capability while also increasing source--learner discrepancy.
\textbf{(c)} Guidance scores for two anchor policies, showing that the preferred temporal distance can shift over training.
Details are provided in Appendix~\ref{app:fig2_details}.}
\label{fig:quality_proximity}
\end{figure*}

\section{Motivation: Why Learn from the Near Future?}
\label{sec:quality_proximity}

Temporal self-distillation introduces a natural question: \textit{how far into the future should the learner look?}
A teacher that is too close may contribute little new capability, while one that is too far may be difficult for the learner to absorb.
We characterize this trade-off through two properties of a guidance source: \textbf{signal quality} $Q$, which captures useful capability beyond the current learner, and \textbf{guidance cost} $V$, which captures the difficulty of transferring that capability back.

Here, $Q$ and $V$ are general conceptual quantities whose concrete realization depends on the transfer mechanism.
In mixed-policy RL, $Q$ can be estimated from additional verified successes, while $V$ is closely related to off-policy variance and can be approximated using source--learner discrepancy measures such as token-level KL divergence, log-ratio variance, or token log-probability gap.

Checkpoint distance $\Delta$ naturally controls this trade-off.
Looking farther ahead gives the policy more opportunity to acquire new capabilities, potentially increasing $Q$, but also more opportunity to diverge from the learner, increasing $V$.
This motivates the \textbf{near-future hypothesis}: the most useful temporal teacher need not be the strongest one, but may lie at an intermediate distance where additional capability remains readily transferable.

Figure~\ref{fig:quality_proximity}(b,c) provides a pilot study along a GRPO training run.
As $\Delta$ increases, later checkpoints provide stronger capability but also become more mismatched with the learner.
Among the candidates examined, the preferred teacher lies at an intermediate temporal distance, and this distance shifts with training stage: from $\Delta=20$ for the early anchor $T=0$ to $\Delta=70$ for the later anchor $T=50$.
This suggests that temporal teacher selection should be learner-dependent rather than fixed.

These observations motivate the methods that follow.
\textbf{NPO} and \textbf{NPD} test whether the near-future preference persists under behavioral and distributional transfer, respectively, while \textbf{AutoNPO} adapts when to intervene and how far along the learning trajectory to look.

\section{Near-Future Policy Optimization}
\label{sec:method}

Having motivated temporal teacher selection, we now present the primary
off-policy realization of temporal self-distillation:
\textbf{Near-Future Policy Optimization (NPO)}.
NPO transfers verified behavior from a nearby future policy to an earlier
learner through trajectory-level mixed-policy RL, while preserving the
underlying RLVR objective. 

\subsection{Off-Policy Trajectory Transfer}
\label{sec:nfpo_mechanism}

Consider an anchor checkpoint $\pi^{(t)}$. Starting from this checkpoint, we
continue training for $\Delta$ steps to obtain a future policy
$\pi_F = \pi^{(t+\Delta)}$. We then restore the anchor checkpoint and initialize
the replayed learner as $\pi_L \leftarrow \pi^{(t)}$. The future policy remains
frozen and serves only as a source of verified guidance behavior during the
replayed segment.

Before replay begins, we roll out $\pi_F$ offline on each prompt $x$, verify
its responses, and retain one correct trajectory $o_x'$ when available
(optionally with length or entropy filtering). Prompts for which the future
policy produces no verified-correct response are excluded from the guidance
cache. The cache is constructed once and reused throughout the NPO segment, so
the future policy does not need to be queried during each replay update.

At each replay step, the current learner samples an on-policy group of $n$
trajectories,
\begin{equation}
    o_i \sim \pi_L(\cdot \mid x),
    \qquad
    i = 1,\ldots,n,
\end{equation}
and we compute the group pass-rate
\begin{equation}
    \hat p(x)
    =
    \frac{1}{n}\sum_{i=1}^{n} r(x,o_i).
\end{equation}
Guidance is injected only when the learner is still struggling with the prompt.
If $\hat p(x) \le \tau_{\mathrm{gate}}$ and a cached trajectory $o_x'$ exists,
we replace one rollout-group slot with the verified future-self trajectory;
otherwise the group remains fully on-policy. Writing the final slot as
$\tilde{o}_n$, the NPO group is
\begin{equation}
\label{eq:nfpo_group}
    \mathcal{G}_{\mathrm{NPO}}(x)
    =
    \{o_1,\ldots,o_{n-1},\tilde{o}_n\},
    \qquad
    \tilde{o}_n =
    \begin{cases}
        o_x' &
        \text{if } \hat p(x) \le \tau_{\mathrm{gate}}
        \text{ and } o_x' \text{ exists},\\
        o_n & \text{otherwise}.
    \end{cases}
\end{equation}

Group-relative advantages are computed over
$\mathcal{G}_{\mathrm{NPO}}(x)$ in the same way as standard GRPO,
\begin{equation}
    A_i
    =
    \frac{
        r_i - \mathrm{mean}(\{r_j\}_{j=1}^{n})
    }{
        \mathrm{std}(\{r_j\}_{j=1}^{n})
    },
\end{equation}
and the learner is updated using the usual clipped objective,
\begin{equation}
\label{eq:nfpo_obj}
\mathcal{L}_{\mathrm{NPO}}(\theta)
=
\mathbb{E}_{x,\mathcal{G}_{\mathrm{NPO}}(x)}
\left[
\frac{1}{n}
\sum_{i=1}^{n}
\frac{1}{|o_i|}
\sum_{\ell=1}^{|o_i|}
\min
\left(
\rho^{q}_{i,\ell}(\theta) A_i,\;
\mathrm{clip}
\big(
\rho^{q}_{i,\ell}(\theta),
1-\epsilon,
1+\epsilon
\big)A_i
\right)
\right],
\end{equation}
where
\begin{equation}
    \rho^{q}_{i,\ell}(\theta)
    =
    \frac{
        \pi_\theta(o_{i,\ell}\mid x,o_{i,<\ell})
    }{
        q_i(o_{i,\ell}\mid x,o_{i,<\ell})
    }.
\end{equation}
For ordinary rollout slots, $q_i$ is the learner policy that generated the
trajectory; for the guidance slot, $q_i=\pi_F$. The formulation therefore
admits the standard source-policy importance correction for the off-policy
guidance trajectory. In practice, Section~\ref{sec:ablation} shows that
omitting this correction has only a marginal effect for the near-future guides
considered here.

NPO therefore changes the \emph{source} of one rollout slot rather than the
learning rule itself: the reward, verifier, group-relative advantage
construction, and clipped RLVR objective remain unchanged. In this sense, NPO
implements temporal self-distillation through off-policy behavioral transfer:
the future policy contributes verified behavior rather than token-level teacher
probabilities.

\subsection{Manual Temporal Interventions}
\label{sec:two_interventions}

We first study two manually chosen intervention points to probe when
near-future transfer is useful. Both use the same mechanism in
Section~\ref{sec:nfpo_mechanism}; they differ only in when the anchor is chosen
and how the future guide is obtained.

\noindent\textbf{Early-Stage Bootstrapping.}
Early in training, many rollout groups contain no correct trajectory and
therefore provide little or no useful group-relative learning signal. We run a
short scout continuation from the base checkpoint, use its final checkpoint as
the future guide, and then restore the base model for replay with NPO.
Because the guide is only slightly ahead, it can expose verified solutions that
the fresh learner does not yet sample reliably while remaining close to the
restarted policy. This accelerates learning through the sparse-reward regime.

\noindent\textbf{Late-Stage Plateau Intervention.}
Later in training, the validation curve may flatten and a stable subset of
prompts can remain unsolved. We continue training beyond the observed plateau,
take a stronger nearby checkpoint as the future guide, and then restore and
replay the plateaued segment with NPO. The future guide contributes
verified-correct trajectories on prompts the earlier learner still fails,
helping the replayed policy escape the observed plateau and reach a higher
final performance level.

\noindent\textbf{Takeaway.}
The same transfer rule addresses sparse early signals and limited later
exploration, motivating adaptive choices of \emph{when} to intervene and
\emph{how far} to roll back.

\begin{figure*}[t]
\centering
\includegraphics[width=0.98\textwidth]{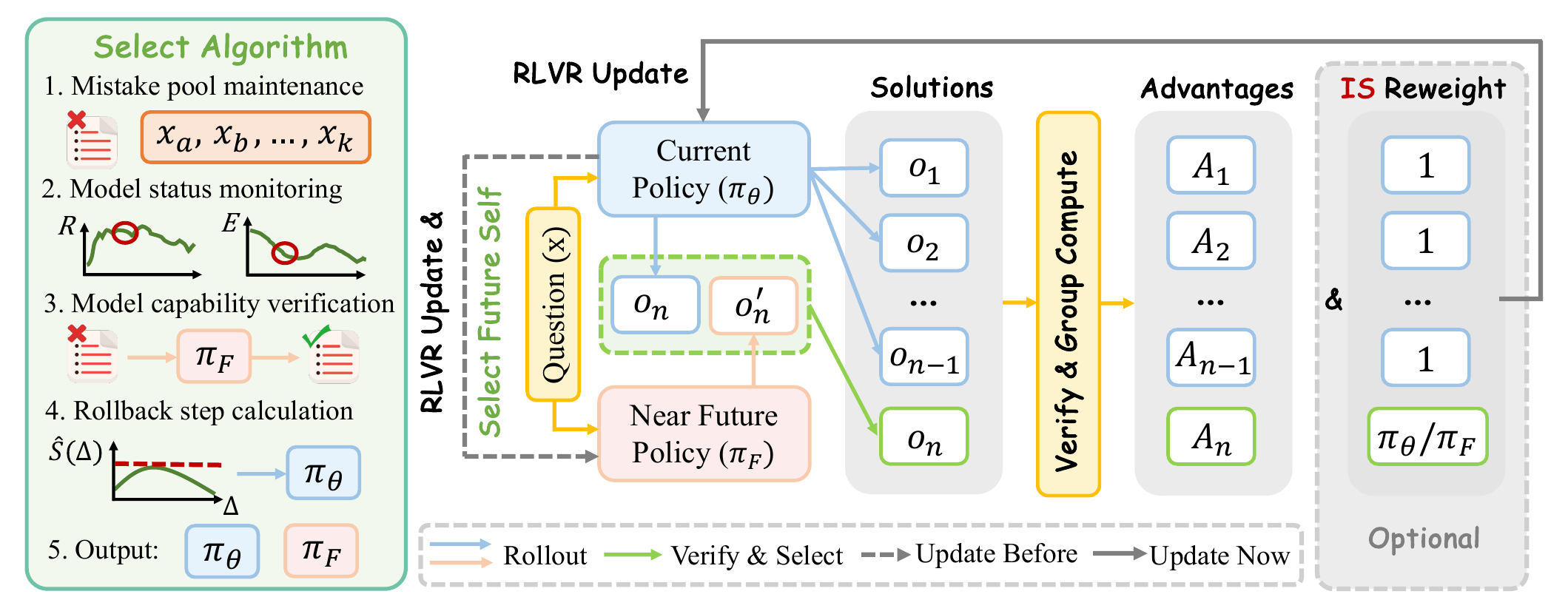}
\caption{
\textbf{Overview of NPO and AutoNPO.}
\textbf{Right:} NPO (\S\ref{sec:nfpo_mechanism}) transfers a verified trajectory
from a near-future policy $\pi_F$ into one slot of the learner's rollout group
when the learner is still struggling with the prompt.
\textbf{Left:} AutoNPO (\S\ref{sec:autonpo}) uses online training signals, a
mistake pool, a capability probe, and an estimated quality--cost score
$\hat{\mathcal{S}}$ to decide \textit{when} to intervene and \textit{how far}
to roll back.
}
\label{fig:method}
\end{figure*}

\subsection{AutoNPO: Adaptive Temporal Selection}
\label{sec:autonpo}

The pilot study in Section~\ref{sec:quality_proximity} suggests that the
appropriate temporal distance is learner-dependent and can change over the
course of training. \textbf{AutoNPO} operationalizes this observation by
adaptively deciding \textit{when} retrospective guidance is useful and
\textit{how far} along the training trajectory to roll back.

To decide \textit{when}, AutoNPO maintains a lightweight mistake pool of
recently failed prompts. Reward stagnation together with declining policy
entropy raises a warning; if this pattern persists, a confirmation probe tests
whether the current policy has acquired new capability on earlier failures.
An intervention is triggered only when the probe indicates that sufficiently
useful guidance has emerged.

To decide \textit{how far}, AutoNPO ranks the available rollback distances
according to the quality--cost trade-off introduced in
Section~\ref{sec:quality_proximity}. We operationalize this trade-off using
\begin{equation}
    \hat{\mathcal{S}}(\Delta)
    =
    \frac{\hat Q(\Delta)}{\hat V(\Delta)},
\end{equation}
where $\hat Q(\Delta)$ estimates the additional verified capability available
from the future guide and $\hat V(\Delta)$ is a KL-based proxy for the guidance
cost induced by source--learner mismatch. AutoNPO selects
\begin{equation}
    \Delta^\star
    =
    \arg\max_{\Delta}
    \hat{\mathcal{S}}(\Delta).
\end{equation}

The current policy then generates verified guidance for the selected failed
prompts, training rolls back to $t-\Delta^\star$, and the segment is replayed
with NPO. Once replay catches up to step $t$, the controller enters a short
cooldown and re-arms, allowing the strengthened learner to continue training
and potentially serve as an even stronger future guide at a later
intervention.

AutoNPO therefore turns a fixed temporal intervention into an adaptive
self-bootstrap process: capability acquired later in training can be reused to
reshape an earlier learner, whose continued optimization may in turn produce
stronger future guidance. Implementation details and full pseudocode are
provided in Appendix~\ref{app:autonpo_algorithm}

\section{Experiments}
\label{sec:experiments}

We organize the experiments around three questions:
(i) whether near-future off-policy transfer improves RLVR across training stages and domains;
(ii) whether the same temporal advantage persists under on-policy distillation; and
(iii) whether the observed gains are consistent with the proposed quality--cost view.

\subsection{Experimental Setup}
\label{sec:setup}

\noindent\textbf{Data and Benchmarks.}
We use the MMFineReason-123K data preparation of
\citet{yang2026selfdistilledrlvr}, a difficulty-filtered subset of
MMFine-Reason-1.8M~\citep{lin2026mmfinereasonclosingmultimodalreasoning}
on which Qwen3-VL-4B-Thinking~\citep{qwen3vl} fails all four rollouts.
We evaluate on eight image-text reasoning benchmarks, six text-only reasoning
benchmarks, and six video reasoning benchmarks; the full benchmark list is
provided in Appendix~\ref{app:exp_details}.
We report accuracy (\%) and the unweighted mean \textit{Avg.}.

\noindent\textbf{Models and Baselines.}
The image-text experiments use Qwen3-VL-8B-Instruct~\citep{qwen3vl}
as the base policy.
For RLVR, we compare against GRPO~\citep{grpo},
LUFFY~\citep{yan2025learning}, ExGRPO~\citep{zhan2025exgrpo}, and
RLEP~\citep{zhang2025rlep}, covering pure on-policy learning,
external-teacher guidance, historical replay, and longer-distance future
guidance.
We also include the distillation baselines OPSD~\citep{zhao2026opsd},
SDPO~\citep{hubotter2026reinforcement}, and
OPD~\citep{agarwal2024policy,lu2025onpolicydistillation}.
OPD uses a frozen Qwen3-VL-32B-Instruct teacher to supervise the student's
own rollouts at the token level.

\noindent\textbf{Implementation Details.}
Our NPO implementation is built on EasyVideoR1~\citep{qin2026easyvideor1easierrlvideo}.
The default GRPO configuration uses an 8192-token context, batch size 256,
$n=8$ rollouts per prompt, sampling temperature $1.0$, learning rate
$1\times10^{-6}$, and clipping thresholds
$\epsilon_{\mathrm{low}}=0.2$ and $\epsilon_{\mathrm{high}}=0.28$.
NPO injects future-self guidance when
$\hat p(x)\le\tau_{\mathrm{gate}}$, with
$\tau_{\mathrm{gate}}=0.6$ by default.
The distillation experiments use token-level Jensen--Shannon divergence
on student-generated responses.
Full training and teacher configurations are given in Appendix~\ref{app:distillation_baselines}.

\subsection{Off-Policy Temporal Self-Distillation: NPO and AutoNPO}
\label{sec:main_results}

\noindent\textbf{Image-text reasoning.}
Table~\ref{tab:image_text} evaluates NPO and AutoNPO on eight image-text
reasoning benchmarks.
Pure on-policy GRPO reaches an average of 60.25.
Applying NPO only in the early stage raises this to 62.12, showing that
near-future transfer is already useful when correct on-policy trajectories
are scarce.
Adding a late-stage intervention further improves the average to 62.84,
while AutoNPO reaches 63.15 without manually selected intervention points.
These results match the two intended roles of NPO:
near-future guidance accelerates early learning and remains useful when
the policy later approaches a plateau.


\begin{table*}[t]
    \caption{%
        Multimodal reasoning results on Qwen3-VL-8B-Instruct (\%, higher is better). OPSD and SDPO use self-distillation; OPD uses a frozen Qwen3-VL-32B-Instruct teacher. Blue rows denote NPO variants. \textbf{Bold} marks the best result per column; \textit{Avg.}\ is the unweighted mean over eight benchmarks.
    }
    \label{tab:image_text}
    \centering
    \begingroup
    \definecolor{TableOneHeader}{rgb}{0.87,0.94,1}
    \definecolor{TableOneHighlight}{rgb}{0.87,0.94,1}
    \definecolor{TableOneInk}{HTML}{245675}
    \resizebox{\linewidth}{!}{%
    \renewcommand{\arraystretch}{1.22}
    \setlength{\tabcolsep}{3.5pt}
    \begin{tabular}{@{}l*{8}{r}@{\hspace{9pt}}r@{}}
        \toprule
        \rowcolor{TableOneHeader}
        \textbf{Method}
            & \textbf{MMMU-Pro} & \textbf{MathVista} & \textbf{MathVision} & \textbf{ZeroBench}
            & \textbf{WeMath} & \textbf{MMBench} & \textbf{MM-Star} & \textbf{MathVerse}
            & \textbf{Avg.} \\
        \midrule
        Qwen3-VL-8B-Instruct
            & 51.75 & 73.80 & 47.37 & 19.76
            & 54.10 & 89.79 & 71.83 & 54.61
            & 57.88 \\
        \midrule
        \multicolumn{10}{@{}l}{\textcolor{TableOneInk}{\textbf{Distillation methods}}} \\[1pt]
        OPSD
            & 53.58 & 75.10 & 47.53 & 21.06
            & 54.95 & 89.70 & 71.57 & 58.03
            & 58.94 \\
        SDPO
            & 52.64 & 74.00 & 47.27 & 25.15
            & 52.19 & 90.03 & 72.21 & 58.44
            & 58.99 \\
        OPD (32B teacher)
            & 53.42 & 76.95 & 48.57 & 21.71
            & 58.19 & 90.15 & 72.00 & 53.67
            & 59.33 \\
        \midrule
        \multicolumn{10}{@{}l}{\textcolor{TableOneInk}{\textbf{RLVR methods}}} \\[1pt]
        GRPO (pure on-policy)
            & 55.78 & 76.20 & 48.82 & 22.60
            & 56.57 & 90.29 & 72.20 & 59.52
            & 60.25 \\
        LUFFY (external teacher)
            & 54.23 & 73.80 & 54.00 & 20.51
            & 52.38 & 89.49 & 69.47 & 55.58
            & 58.68 \\
        ExGRPO (historical replay)
            & 55.49 & 77.30 & 55.46 & 19.01
            & 62.67 & 90.44 & 72.00 & 56.89
            & 61.16 \\
        RLEP (far future)
            & 55.38 & 78.50 & 54.23 & 19.61
            & 62.48 & 90.45 & 72.27 & 58.91
            & 61.48 \\
        \addlinespace[2pt]
        \rowcolor{TableOneHighlight}
        NPO (early only)
            & 56.85 & 76.60 & 54.31 & \textbf{26.35}
            & 62.76 & 90.41 & 70.30 & 59.38
            & 62.12 \\
        \rowcolor{TableOneHighlight}
        NPO (early $+$ late)
            & 57.07 & 76.30 & 54.61 & 24.85
            & \textbf{66.95} & 90.30 & \textbf{72.67} & \textbf{60.00}
            & 62.84 \\
        \rowcolor{TableOneHighlight}
        \textbf{AutoNPO}
            & \textbf{57.24} & \textbf{79.20} & \textbf{55.72} & 24.70
            & 66.00 & \textbf{90.63} &72.63 & 59.11
            & \textbf{63.15} \\
        \bottomrule
    \end{tabular}
    }
    \endgroup
\end{table*}

\begin{table}[t]
\caption{\textbf{Text (top) and video (bottom) reasoning.} Text methods continue from the same LUFFY policy. \textbf{Bold} marks the better score; \textit{Avg.}\ is computed within each domain. Full comparisons are in Appendix~\ref{app:extension_results}.}
\label{tab:text_ext}
\label{tab:extensions}
\centering
\begingroup
\definecolor{ExtensionBlue}{rgb}{0.87,0.94,1}
\fontsize{8}{10}\selectfont
\renewcommand{\arraystretch}{1.05}
\setlength{\tabcolsep}{4.5pt}
\resizebox{\linewidth}{!}{%
\begin{tabular}{@{}l*{7}{r}@{}}
\toprule
\textbf{Method} & \textbf{AIME 2024} & \textbf{AIME 2025} & \textbf{AMC} & \textbf{MATH-500} & \textbf{Minerva Math} & \textbf{OlympiadBench} & \textbf{Avg.}\\
\midrule
GRPO & 24.8 & 17.8 & \textbf{67.5} & 88.4 & 38.6 & 55.3 & 48.7\\
\rowcolor{ExtensionBlue}
NPO & \textbf{34.6} & \textbf{25.7} & 66.9 & \textbf{89.2} & \textbf{40.4} & \textbf{58.2} & \textbf{52.5}\\
\midrule
\textbf{Method} & \textbf{TempCompass} & \textbf{MotionBench} & \textbf{MLVU-Dev} & \textbf{VideoMathQA} & \textbf{Charades-STA} & \textbf{VideoMME-v2} & \textbf{Avg.}\\
\midrule
GRPO & 75.98 & 57.01 & 75.80 & 50.33 & 57.65 & 33.03 & 58.30\\
\rowcolor{ExtensionBlue}
NPO & \textbf{76.47} & \textbf{59.17} & \textbf{76.40} & \textbf{52.71} & \textbf{58.06} & \textbf{33.81} & \textbf{59.44}\\
\bottomrule
\end{tabular}}
\endgroup
\end{table}

\noindent\textbf{Text-only and video reasoning.}
We next test whether the same off-policy temporal transfer generalizes
beyond image-text reasoning (Table~\ref{tab:extensions}).
For text reasoning, we follow the experimental settings of
LUFFY~\citep{yan2025learning} and ExGRPO~\citep{zhan2025exgrpo} to ensure comparability.
Starting from the same LUFFY policy, NPO reaches an average of 52.5,
compared with 48.7 for on-policy GRPO, and also exceeds continual LUFFY
and ExGRPO in the full comparison reported in
Appendix~\ref{app:extension_results}.
For video reasoning, NPO improves over GRPO on all six benchmarks,
raising the average from 58.30 to 59.44.
The gains across text, image-text, and video reasoning indicate that the
benefit is not specific to a single modality or benchmark family.

\subsection{On-Policy Temporal Self-Distillation: NPD}
\label{sec:npd}

\begin{wrapfigure}{R}{0.415\textwidth}
\centering
\includegraphics[width=\linewidth]{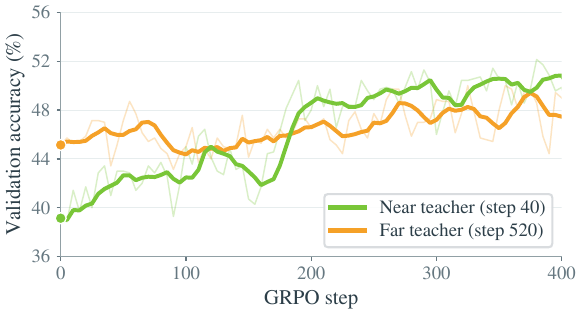}
\captionsetup{font=small,skip=4pt}
\caption{
\textbf{Continued RL after NPD.}
Thin lines show raw evaluations; thick lines show trailing means over five evaluations.
}
\label{fig:opd_extension}
\end{wrapfigure}

We next ask whether the near-future advantage depends on off-policy
trajectory transfer.
To isolate this question, we study an on-policy realization of temporal
self-distillation, which we call
\textbf{Near-Future Policy Distillation (NPD)}.
NPD uses the standard on-policy distillation objective but selects a
near-future checkpoint as the teacher.
Unlike NPO, where the future policy generates the transferred trajectory,
NPD evaluates the teacher on student-generated responses and transfers
token-level distributions on the learner's own rollout distribution.

We compare a near-future teacher with a far-future teacher while keeping
the student initialization, distillation objective, and subsequent GRPO
stage fixed.
Both teachers are frozen checkpoints from the same GRPO training process,
and each distills its behavior into the original Qwen3-VL-8B-Instruct
student.
We then select the best checkpoint from each distillation run and continue
both students using the same GRPO configuration.
Implementation details are given in Appendix~\ref{app:opd_extension}.

The near-future teacher leads to stronger downstream learning.
After subsequent RL, the near-guided student reaches an eight-benchmark
average of 63.23, compared with 61.92 for the far-guided student, and
performs better on seven of the eight benchmarks
(Table~\ref{tab:opd_extension}).
Interestingly, the near-guided student starts the subsequent RL stage from
a lower validation score, yet improves substantially more during continued
training (Figure~\ref{fig:opd_extension}).

Because NPD applies teacher supervision on the learner's own rollout
distribution, this advantage cannot be attributed solely to reduced
off-policy trajectory shift.
More importantly, the result shows that higher immediate post-distillation
performance does not necessarily imply better learnability under subsequent
RL: the near-future teacher produces a student that starts weaker, but learns more effectively afterward.

\subsection{Analysis and Discussion}
\label{sec:analysis}

\begin{figure}[ht]
    \centering
    \includegraphics[width=\textwidth]{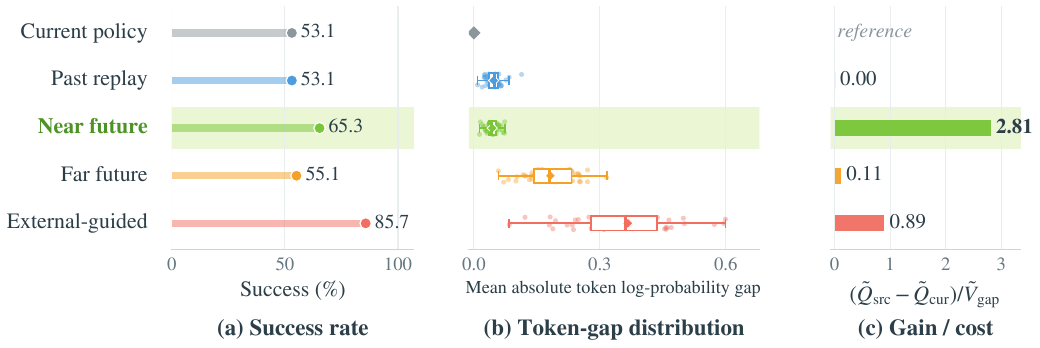}
    \caption{
    \textbf{Source-level evidence for the quality--cost trade-off.}
    \textbf{(a)} Verified success $\tilde Q$ across guidance sources.
    \textbf{(b)} Source--learner mismatch measured by mean absolute token
    log-probability gap $\tilde V_{\mathrm{gap}}$.
    \textbf{(c)} Incremental verified success per unit mismatch cost.
    Near-future guidance provides the largest measured usable-gain ratio among the compared sources.}
    \label{fig:source_tradeoff}
\end{figure}

\paragraph{Source-level evidence for the quality--cost view.}
Figure~\ref{fig:source_tradeoff} compares several guidance sources in terms
of verified success and source--learner mismatch.
Current-policy and historical-replay trajectories remain close to the
learner but add little new success, whereas the externally guided policy provides
substantially higher raw success at a much larger token-level discrepancy.
The near-future source occupies an intermediate regime: its verified
success rate rises from 53.1\% for the current policy to 65.3\%, while its
token-level gap remains close to historical replay.
By contrast, the longer-distance future source incurs greater discrepancy
without a commensurate increase in success.
The resulting incremental gain-per-cost proxy is highest for the
near-future source (2.81), compared with 0.89 for the externally guided policy and
0.11 for the longer-distance future source.
This source-level evidence is consistent with the quality--cost hypothesis
in Section~\ref{sec:quality_proximity}: the most useful guide need not be
either the closest or the strongest one.

The same qualitative pattern is observed under two different mismatch
proxies.
Figure~\ref{fig:quality_proximity} uses a KL-based cost proxy along the
checkpoint sequence, whereas Figure~\ref{fig:source_tradeoff} measures
absolute token log-probability gaps across source types.
Together, these analyses support treating source--learner mismatch as a
general guidance cost rather than tying it to a single metric.

\begin{figure*}[ht] 
\centering 
\includegraphics[width=1\textwidth]
{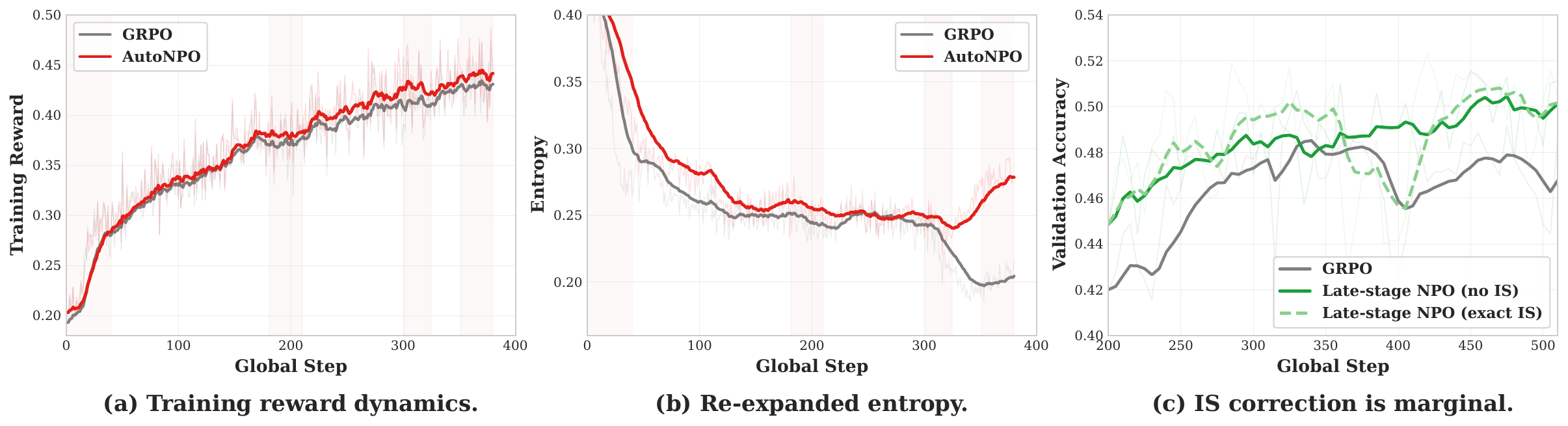} 
\caption{Training dynamics of (Auto)NPO compared with vanilla GRPO. \textbf{\textcolor{red}{Red shaded regions}} in panels (a) and (b) highlight AutoNPO's intervention windows. \textbf{(a)}~Training reward dynamics: AutoNPO stays above GRPO throughout, with the gap widening step-wise after each intervention window. \textbf{(b)}~Generation entropy: GRPO undergoes a steady entropy collapse, while AutoNPO's interventions actively re-expand exploration and preserve a substantially higher-entropy policy. \textbf{(c)}~IS-correction ablation on the late-stage intervention: the two NPO variants with and without exact importance-sampling correction both clearly outperform GRPO, indicating that IS correction is marginal for NPO. Thick lines are smoothed curves; thin lines show raw logged values.}\label{fig:training_dynamics} \end{figure*}

\paragraph{Training dynamics.}
Figure~\ref{fig:training_dynamics} shows how AutoNPO changes the training
process through localized retrospective interventions.
AutoNPO improves more rapidly in the early stage and remains above GRPO
throughout training; later intervention windows produce additional
step-wise improvements rather than a uniform shift across the entire run.
At the same time, GRPO's generation entropy steadily declines, consistent
with a narrowing rollout distribution, whereas AutoNPO maintains higher
entropy and re-expands after intervention windows.
This behavior suggests that retrospective transfer does more than replay
successful answers: it can reopen useful regions of the learner's rollout
distribution for subsequent on-policy exploration.

The late-stage comparison in Figure~\ref{fig:training_dynamics} also shows
that the NPO gain is largely preserved when the source-policy importance
correction is omitted.
We therefore retain the correction in the formal off-policy formulation,
but do not treat it as a critical ingredient of near-future transfer.

\paragraph{Ablations.}
\label{sec:ablation}
The ablations in Appendix~\ref{app:ablation_discussion}
(Table~\ref{tab:ablation_compact}) test the main design choices behind
adaptive temporal transfer.
Restricting the rollback search range or fixing a single temporal distance
reduces performance relative to AutoNPO, consistent with the
learner-dependent teacher selection motivated in
Section~\ref{sec:quality_proximity}.
Guidance gating is also important: overly restrictive gating misses useful
transfer opportunities, whereas overly permissive gating replaces too much
on-policy experience.
Finally, omitting the source-policy importance correction changes
performance only marginally for the near-future guides considered here.

\par\smallskip
\noindent
\begin{minipage}[t]{0.515\textwidth}
\vspace{0pt}
\raggedright
\phantomsection\label{sec:analysis_cost}
\textbf{Cost analysis.}\quad
To separate algorithmic gains from simply spending more rollout compute,
we compare AutoNPO and GRPO under the same 500 GPU-hour budget, including
AutoNPO's rollback and guidance costs.
AutoNPO reaches 53.43\% best validation accuracy, compared with 50.71\%
for GRPO with $n=8$ and 51.14\% for GRPO with twice as many rollouts
($n=16$).
Their corresponding eight-benchmark averages are 63.15, 60.25, and 61.14.
Thus, the AutoNPO advantage remains even when the on-policy baseline is
given substantially more rollout capacity.
Full benchmark results are reported in
Appendix~\ref{app:compute_comparison}.
\end{minipage}\hfill
\begin{minipage}[t]{0.46\textwidth}
\vspace{0pt}
\centering
\includegraphics[width=\linewidth]{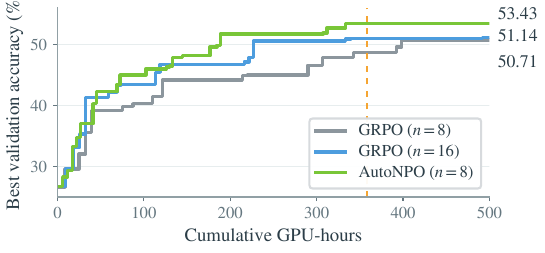}
\captionsetup{font=small,skip=4pt}
\captionof{figure}{Best validation accuracy within 500 GPU-hours. The dashed line marks AutoNPO's first intervention.}
\label{fig:compute_comparison}
\end{minipage}\par

\section{Related Work}
\label{sec:related}

\paragraph{RLVR and trajectory-source methods.}
GSPO and SAPO refine the on-policy RLVR update~\citep{zheng2025gspo,gao2025sapo}. LUFFY adds external teacher traces~\citep{yan2025learning}, while ExGRPO and RLEP reuse historical or stronger-policy experience~\citep{zhan2025exgrpo,zhang2025rlep}. NPO selects a nearby later checkpoint to balance additional capability against source--learner mismatch.

\paragraph{Self-distillation and self-guided learning.}
ReST and STaR learn from successful model generations~\citep{gulcehre2023reinforced,zelikman2022star}. Context distillation uses privileged context~\citep{snell2022learning}, while on-policy distillation supplies teacher predictions on student-generated responses~\citep{agarwal2024policy,lu2025onpolicydistillation}. OPSD and SDPO condition these predictions on reference solutions or feedback~\citep{zhao2026opsd,hubotter2026reinforcement}. Temporal self-distillation asks \emph{which later checkpoint should teach an earlier learner}. We study this choice through NPO's off-policy trajectories and NPD's on-policy token targets. Further comparisons appear in Appendix~\ref{app:related_work}.

\section{Conclusion and Future Work}
\label{sec:conclusion}

We introduce \textbf{temporal self-distillation}: a policy can learn from a stronger later version of itself, provided the teacher is sufficiently capable yet still close enough to be readily learned from. We capture this trade-off through $\mathcal{S}=Q/V$ and instantiate it through two complementary mechanisms: \textbf{NPO} for off-policy behavioral transfer and \textbf{NPD} for on-policy distributional transfer. Across image-text, text-only, and video reasoning, near-future guidance consistently improves learning, and its advantage persists across both transfer settings.

More broadly, \textbf{AutoNPO} turns temporal self-distillation into an adaptive self-bootstrap process by deciding when to intervene and how far back to revisit. A natural next step is to let sufficiently capable models reason over their own learning histories and decide \emph{when, where, and how} to improve themselves, moving toward more self-directed forms of recursive self-improvement.
\subsection*{AI use statement}
\label{sec:ai_use}

Generative AI tools assisted with manuscript editing, analysis of reviewer feedback, planning of follow-up experiments, and \LaTeX{} formatting. The authors are responsible for the scientific claims, experimental results, and final content of the paper.

\clearpage
\bibliographystyle{unsrtnat}
\bibliography{references}

\clearpage
\appendix

\section{Additional Experimental Results}
\label{app:additional_results}

\subsection{Experimental Details}
\label{app:exp_details}

\paragraph{Training data.}
We use the MMFineReason-123K data preparation of \citet{yang2026selfdistilledrlvr}, derived from the MMFineReason-1.8M corpus~\citep{lin2026mmfinereasonclosingmultimodalreasoning} via difficulty-based filtering.
Each training sample is rolled out four times with Qwen3-VL-4B-Thinking~\citep{qwen3vl}, and only samples on which the model fails every attempt are kept.
This criterion removes trivial examples and concentrates training on problems where improvement is still possible, matching the regime where NPO is designed to help.

\paragraph{Image-text benchmarks.}
For evaluation, we use eight multimodal reasoning benchmarks that cover visually grounded mathematics, broad subject knowledge, and fine-grained visual understanding.
MathVista~\citep{lumathvista}, MathVision~\citep{wang2024measuring}, WeMath~\citep{qiao2025we}, and MathVerse~\citep{zhang2024mathverse} target mathematical reasoning in visual contexts, from general problem solving to competition-level diagrams and structured difficulty tiers.
MMMU-Pro~\citep{yue2025mmmu} and MMBench~\citep{liu2024mmbench} test broad multi-discipline knowledge and general multimodal competence.
MM-Star~\citep{chen2024we} stresses fine-grained visual discrimination, and ZeroBench~\citep{roberts2025zerobench} serves as a hard stress test for the tail of the reasoning distribution.

\paragraph{Models and baselines.}
All image-text experiments use Qwen3-VL-8B-Instruct~\citep{qwen3vl} as the base policy. NPO and the trajectory-based RL baselines use a GRPO-style group-based RLVR backbone.
GRPO~\citep{grpo} is the pure on-policy reference.
LUFFY~\citep{yan2025learning} imports trajectories from a stronger external teacher, representing the high-quality but high-shift regime.
ExGRPO~\citep{zhan2025exgrpo} reuses successful trajectories from historical replay buffers, and RLEP~\citep{zhang2025rlep} replays trajectories from a far-future model.
We also report the Base LLM without post-training as a reference point for overall RL gains.
The distillation baselines are OPSD~\citep{zhao2026opsd}, SDPO~\citep{hubotter2026reinforcement}, and OPD~\citep{agarwal2024policy,lu2025onpolicydistillation}; their implementations are described below.

\paragraph{Text and video extensions.}
For text-only reasoning, we follow the matched continuation setting of prior mixed-policy RLVR work~\citep{yan2025learning,zhan2025exgrpo}.
All methods continue from the same LUFFY policy; the training data, verifier, and evaluation protocol are held fixed.
The six text-only benchmarks are AIME 2024, AIME 2025, AMC~\citep{li2024numinamath}, MATH-500~\citep{hendrycks2021measuringmathematicalproblemsolving}, Minerva Math~\citep{dataset_minerva}, and OlympiadBench~\citep{dataset_olympiad}.
The six video benchmarks are TempCompass~\citep{liu2024tempcompassvideollmsreally}, MotionBench~\citep{hong2025motionbenchbenchmarkingimprovingfinegrained}, MLVU-Dev~\citep{zhou2025mlvubenchmarkingmultitasklong}, VideoMathQA~\citep{rasheed2025videomathqabenchmarkingmathematicalreasoning}, Charades-STA~\citep{gao2017talltemporalactivitylocalization}, and VideoMME-v2~\citep{fu2026videommev2stagebenchmarkscomprehensive}.
These benchmarks cover temporal comparison, motion understanding, long-video QA, video-grounded mathematics, temporal localization, and general video QA.

\paragraph{GRPO-based implementation details.}
Our NPO implementation is built on the EasyVideoR1~\citep{qin2026easyvideor1easierrlvideo} framework.
The default GRPO-based configuration uses an 8192-token context, split into a prompt budget of 4096 and a response budget of 4096, at training and evaluation time.
The default learning rate is $1 \times 10^{-6}$ with a batch size of 256, and each prompt is sampled $n=8$ times at temperature $1.0$.
Clipping thresholds are set to $\epsilon_{\text{low}}=0.2$ and $\epsilon_{\text{high}}=0.28$, and we omit both the KL penalty and entropy regularization from the objective.
For all NPO variants, the guidance gate in Section~\ref{sec:nfpo_mechanism} uses $\tau_{\mathrm{gate}}=0.6$, implemented as \texttt{mix\_policy\_accuracy\_threshold}; a future-self trajectory is injected only when $\hat p(x)\le\tau_{\mathrm{gate}}$, leaving confident groups untouched.
The GRPO-based experiments use 4 compute nodes, each with 8 NVIDIA H200 140GB GPUs.

\paragraph{Distillation baselines.}
\label{app:distillation_baselines}
The three distillation baselines learn from token-level teacher distributions evaluated along student-generated responses. We use Jensen--Shannon divergence over the teacher's top-256 tokens, with the remaining probability mass collected into a tail category. Teacher targets are detached from the gradient, and the update contains no additional outcome-reward PPO loss. For OPSD and SDPO, we follow the experimental setting of \citet{yang2026selfdistilledrlvr}: MMFineReason-123K, batch size 256, a rollout group size of eight, a learning rate of $1\times10^{-5}$, and maximum prompt and response lengths of 4,096 tokens each. The sampling temperature is $1.0$ for all three baselines; the external-teacher OPD baseline uses a learning rate of $1\times10^{-6}$.

\textbf{OPSD.} Following the multimodal reference-context setup of \citet{yang2026selfdistilledrlvr}, the self-teacher receives the question, image, and a reference solution from the MMFineReason data. These reference traces are generated by Qwen3-VL-235B-A22B-Thinking and verified as correct. Our implementation uses a frozen copy of the initial 8B model as the teacher and supplies the reference trace directly, without a separate hint-generation stage. When a reference trace is unavailable, the implementation falls back to the final ground-truth answer. The student generates its response from the original question and image, without the reference context.

\textbf{SDPO.} The current student also serves as the self-teacher, conditioned on feedback from its own sampled attempts. For each prompt, the implementation samples eight candidate responses, verifies them, and selects the shortest correct and shortest incorrect response when available as teacher-side context. A separate student rollout provides the response used for distillation. Missing feedback components are left empty. The teacher follows the current student parameters, with stop-gradient applied during teacher scoring; dataset reference traces are not supplied to this teacher.

\textbf{OPD.} A frozen Qwen3-VL-32B-Instruct model supervises the 8B student on the same image, question, and student-generated prefixes. The teacher receives no additional reference solution. The configuration uses a batch size of 256, one student rollout per prompt, and maximum prompt and response lengths of 16,384 and 4,096 tokens, respectively. Table~\ref{tab:image_text} reports the distilled student directly, without subsequent RL training.

\subsection{Full Text and Video Results}
\label{app:extension_results}

Table~\ref{tab:extensions_full} retains the complete continuation comparison for text reasoning and the base-model reference for video reasoning. In the text experiment, GRPO denotes on-policy continuation from the LUFFY starting policy. AIME is the American Invitational Mathematics Examination, and AMC is the American Mathematics Competitions; Minerva Math and OlympiadBench are written out in Table~\ref{tab:extensions}.

\begin{table}[t]
\centering
\begingroup
\definecolor{ExtensionFullBlue}{rgb}{0.87,0.94,1}
\caption{\textbf{Full text and video reasoning results.} (a) Text continuation methods start from the same LUFFY policy; GRPO is the on-policy continuation baseline. (b) Video results cover six benchmarks; gain is the difference between NPO and GRPO. Averages are computed within each panel.}
\label{tab:extensions_full}
{\small\textbf{(a) Text reasoning}}\par\smallskip
\resizebox{\linewidth}{!}{%
\begin{tabular}{lccccccc}
\toprule
\rowcolor{TableHeader}
\textbf{Method} & \textbf{AIME 2024} & \textbf{AIME 2025} & \textbf{AMC} & \textbf{MATH-500} & \textbf{Minerva Math} & \textbf{OlympiadBench} & \textbf{Avg.}\\
\midrule
LUFFY & 29.4 & 23.1 & 65.6 & 87.6 & 37.5 & 57.2 & 50.1\\
$\hookrightarrow$ Continual LUFFY & 30.7 & 22.5 & 66.2 & 86.8 & \textbf{41.2} & 55.3 & 50.4\\
$\hookrightarrow$ GRPO & 24.8 & 17.8 & \textbf{67.5} & 88.4 & 38.6 & 55.3 & 48.7\\
$\hookrightarrow$ ExGRPO & 32.3 & \textbf{25.7} & 65.6 & 87.6 & 40.1 & 57.0 & 51.4\\
\rowcolor{ExtensionFullBlue}
$\hookrightarrow$ \textbf{NPO} & \textbf{34.6} & \textbf{25.7} & 66.9 & \textbf{89.2} & 40.4 & \textbf{58.2} & \textbf{52.5}\\
\bottomrule
\end{tabular}}
\par\medskip
{\small\textbf{(b) Video reasoning}}\par\smallskip
{\small
\begin{tabularx}{\linewidth}{l*{4}{>{\raggedleft\arraybackslash}X}}
\toprule
\rowcolor{TableHeader}
\textbf{Benchmark} & \textbf{Base} & \textbf{GRPO} & \textbf{NPO} & \textbf{Gain}\\
\midrule
TempCompass & 75.69 & 75.98 & \textbf{76.47} & +0.49\\
MotionBench & 56.14 & 57.01 & \textbf{59.17} & +2.16\\
MLVU-Dev & 73.32 & 75.80 & \textbf{76.40} & +0.60\\
VideoMathQA & 47.81 & 50.33 & \textbf{52.71} & +2.38\\
Charades-STA & 56.00 & 57.65 & \textbf{58.06} & +0.41\\
VideoMME-v2 & 32.50 & 33.03 & \textbf{33.81} & +0.78\\
\midrule
\rowcolor{TableHeader}
\textbf{Avg.} & 56.91 & 58.30 & \textbf{59.44} & \textbf{+1.14}\\
\bottomrule
\end{tabularx}}
\endgroup
\end{table}

\subsection{Ablation Discussion}
\label{app:ablation_discussion}

\begin{table*}[t]
    \centering
    \caption{Ablation study on AutoNPO design choices. \textit{Avg.}\ is the unweighted mean over the eight multimodal reasoning benchmarks.}
    \label{tab:ablation_compact}
    \resizebox{\linewidth}{!}{%
    \renewcommand{\arraystretch}{1.15}
    \setlength{\tabcolsep}{3pt}
    \begin{tabular}{lcccccccc|c}
        \toprule
        \rowcolor{TableHeader}
        \textbf{Method}
            & \textbf{MMMU-Pro} & \textbf{MathVista} & \textbf{MathVision} & \textbf{ZeroBench}
            & \textbf{WeMath} & \textbf{MMBench} & \textbf{MM-Star} & \textbf{MathVerse}
            & \textbf{Avg.} \\
        \midrule
        GRPO
            & 55.78 & 76.20 & 48.82 & 22.60 & 56.57 & 90.29 & 72.20 & 59.52 & 60.25 \\
        \midrule
        \rowcolor{TableHighlight}
        NPO, early-stage only
            & 56.85 & 76.60 & 54.31 & \textbf{26.35} & 62.76 & 90.41 & 70.30 & 59.38 & 62.12 \\
        \rowcolor{TableHighlight}
        NPO, early $+$ late-stage
            & 57.07 & 76.30 & 54.61 & 24.85 & \textbf{66.95} & 90.30 & 72.67 & 60.00 & 62.84 \\
        \rowcolor{TableHighlight}
        \textbf{AutoNPO}
            & \textbf{57.24} & \textbf{79.20} & \textbf{55.72} & 24.70 & 66.00 & 90.63 & 72.63 & 59.11 & \textbf{63.15} \\
        \midrule
        Search range $\Delta_{\max}=20$
            & 57.09 & 77.90 & 53.29 & 23.95 & 67.24 & 90.60 & 72.53 & 59.90 & 62.81 \\
        Search range $\Delta_{\max}=80$
            & 55.98 & 76.60 & 52.58 & 23.05 & 65.62 & 90.83 & 72.07 & 59.26 & 62.00 \\
        Fixed $\Delta=40$
            & 56.97 & 76.85 & 55.39 & 25.60 & 63.24 & \textbf{90.89} & 72.77 & \textbf{60.15} & 62.73 \\
        \midrule
        Gate $\tau_{\mathrm{gate}}=0.3$
            & 55.38 & 77.75 & 52.99 & 23.50 & 60.86 & 90.45 & 72.53 & 58.87 & 61.54 \\
        Gate $\tau_{\mathrm{gate}}=0.8$
            & 56.38 & 76.75 & 54.62 & 24.10 & 64.19 & 90.08 & \textbf{73.10} & 56.92 & 62.02 \\
        \midrule
        Late-stage NPO w/o exact IS
            & 56.50 & 76.60 & 55.03 & 24.40 & 66.10 & 90.30 & 71.90 & 59.99 & 62.60 \\
        \bottomrule
    \end{tabular}%
    }
\end{table*}

Table~\ref{tab:ablation_compact} separates four design choices in AutoNPO.
The manual interventions isolate stage effects: early guidance improves the average from 60.25 to 62.12 and is strongest on ZeroBench, where correct on-policy samples are scarce, while late-stage guidance reaches 62.84 and gives the largest gains on WeMath, where the vanilla run plateaus.
AutoNPO combines these effects without manual scheduling and reaches 63.15.

The controller ablations show that the gain is not explained by simply injecting more off-policy trajectories.
Changing the rollback search range from $\Delta_{\max}=20$ to $80$ weakens performance at longer distances, consistent with the near-future trade-off.
The gate sweep changes the default $\tau_{\mathrm{gate}}=0.6$ to 0.3 or 0.8: the lower gate misses many learnable prompts, while the higher gate replaces too many on-policy samples.
Exact importance-sampling correction is also empirically marginal for near-future guidance: late-stage NPO reaches 62.84 with exact IS and 62.60 without it, and the training curves in Figure~\ref{fig:training_dynamics}(c) remain close.
Dropping IS saves memory and compute because guidance log-probabilities no longer need to be stored or recomputed; this simplification should not be read as a general claim for far-off-policy sources.

\subsection{Distillation Extension}
\label{app:opd_extension}

\paragraph{OPD and GRPO settings.}
Both students start from Qwen3-VL-8B-Instruct. The frozen teachers are the step-40 and step-520 checkpoints of the vanilla GRPO run. OPD uses MMFineReason data, a batch size of 256, one student rollout per prompt, temperature 1.0, and learning rate $3\times10^{-7}$. The loss is token-level Jensen--Shannon divergence over the teacher's top-256 tokens, with the remaining probability mass grouped into a tail category; no outcome-reward PPO loss is applied during OPD. The best checkpoints from the near- and far-teacher OPD runs are selected for subsequent RL, at OPD steps 95 and 165, respectively. Both then use the same GRPO data and configuration: batch size 256, eight rollouts per prompt, temperature 1.0, learning rate $10^{-6}$, clipping thresholds 0.2 and 0.28, and a maximum of 500 updates. The data order is fixed, and both runs use seed 1. Both stages allow 16,384 prompt tokens and 4,096 response tokens. The OPD durations differ; the matched configuration applies to the subsequent GRPO stage.

\begin{table}[t]
\centering
\caption{\textbf{Benchmark results for the distillation extension.} All students start from Qwen3-VL-8B-Instruct. The step-40 and step-520 teachers are frozen 8B GRPO checkpoints. The 32B-teacher row reports pure OPD without subsequent RL and is included as a reference. \textbf{Bold} marks the best result per column, and \textit{Avg.}\ is the unweighted mean across eight benchmarks.}
\label{tab:opd_extension}
\begingroup
\definecolor{OPDNearHighlight}{rgb}{0.87,0.94,1}
\renewcommand{\arraystretch}{1.1}
\setlength{\tabcolsep}{3.5pt}
\resizebox{\linewidth}{!}{%
\begin{tabular}{@{}l*{8}{r}@{\hspace{9pt}}r@{}}
\toprule
\textbf{Method} & \textbf{MMMU-Pro} & \textbf{MathVista} & \textbf{MathVision} & \textbf{ZeroBench} & \textbf{WeMath} & \textbf{MMBench} & \textbf{MM-Star} & \textbf{MathVerse} & \textbf{Avg.}\\
\midrule
Qwen3-VL-8B-Instruct & 51.75 & 73.80 & 47.37 & 19.76 & 54.10 & 89.79 & 71.83 & 54.61 & 57.88\\
GRPO & 55.78 & 76.20 & 48.82 & 22.60 & 56.57 & 90.29 & 72.20 & 59.52 & 60.25\\
OPD (32B teacher) & 53.42 & \textbf{76.95} & 48.57 & 21.71 & 58.19 & 90.15 & 72.00 & 53.67 & 59.33\\
\midrule
OPD (far, step 520) $+$ GRPO & 56.50 & 75.40 & 54.84 & \textbf{25.15} & 62.57 & 90.49 & 71.67 & 58.74 & 61.92\\
\rowcolor{OPDNearHighlight}
OPD (near, step 40) $+$ GRPO & \textbf{58.57} & 76.15 & \textbf{56.64} & 24.70 & \textbf{65.24} & \textbf{90.96} & \textbf{72.77} & \textbf{60.78} & \textbf{63.23}\\
\bottomrule
\end{tabular}}
\endgroup
\end{table}

\paragraph{Validation curves.}
Figure~\ref{fig:opd_extension} reports accuracy on 700 validation prompts, evaluated every five updates with one response per prompt, temperature 0.7, top-$p$ 0.8, top-$k$ 20, and presence penalty 1.5. Before RL, the near- and far-guided students score 39.14\% and 45.14\%; their observed peaks are 52.14\% at step 380 and 51.43\% at step 435, giving improvements of 13.00 and 6.29 percentage points. These validation scores are separate from the benchmark averages in Table~\ref{tab:opd_extension}. The figure shows steps 0--400 with trailing means over up to five evaluations; peaks use all available evaluations, through steps 500 and 490 for the near and far runs, respectively.

\subsection{Compute Comparison}
\label{app:compute_comparison}

\begin{table}[t]
\caption{Benchmark results for the compute comparison (\%). Bold marks the best result; Avg. is the mean over eight benchmarks.}
\label{tab:compute_benchmarks}
\centering
\begingroup
\definecolor{ComputeHighlight}{HTML}{EFF8F1}
\setlength{\tabcolsep}{3.5pt}
\renewcommand{\arraystretch}{1.15}
\resizebox{\linewidth}{!}{%
\begin{tabular}{@{}lrrrrrrrrr@{}}
\toprule
Method & MMMU-Pro & MathVista & MathVision & ZeroBench & WeMath & MMBench & MM-Star & MathVerse & Avg. \\
\midrule
GRPO ($n=8$) & 55.78 & 76.20 & 48.82 & 22.60 & 56.57 & 90.29 & 72.20 & \textbf{59.52} & 60.25 \\
GRPO ($n=16$) & 55.67 & 74.75 & 53.78 & 23.50 & 62.19 & 90.16 & 69.57 & 59.50 & 61.14 \\
\rowcolor{ComputeHighlight}
AutoNPO & \textbf{57.24} & \textbf{79.20} & \textbf{55.72} & \textbf{24.70} & \textbf{66.00} & \textbf{90.63} & \textbf{72.63} & 59.11 & \textbf{63.15} \\
\bottomrule
\end{tabular}}
\endgroup
\end{table}

\subsection{Figure Details}
\label{app:fig2_details}

\paragraph{Figure~\ref{fig:quality_proximity}.}
Panel (a) is a conceptual illustration of the quality--cost trade-off. Dots indicate source quality and policy proximity, shaded disks represent guidance cost, and triangles illustrate the useful signal after accounting for that cost. Their positions and sizes are schematic rather than measured comparisons between methods.

Panels (b,c) use checkpoints from a vanilla GRPO run with Qwen3-VL-8B-Instruct. For an anchor at step $T$, the quality proxy $\widetilde Q(\Delta)$ is the positive validation-score gain over that anchor, using the logged \texttt{val.reward\_score}. The validation sequence is smoothed with an eight-observation window, and the gains with a three-observation window. Panel (b) then takes a cumulative maximum to display the gain envelope. We use validation improvement here as a proxy for additional capability.

The cost proxy is $\widetilde V(\Delta)=\exp((\Delta))$, where $d_T(\Delta)$ is the empirical token KL estimated on trajectories generated by the anchor and scored by later checkpoints. The exponential transform sets the scale of the heuristic cost; it is not a direct measurement of gradient variance. Panel (b) uses anchor $T=0$. Panel (c) plots $\widetilde Q/\widetilde V$ for anchors $T=0$ and $T=50$, using the smoothed gains without the cumulative-maximum operation. The marked points are the highest scores among the sampled distances, at 20 and 70 steps, respectively.

\section{Additional Related Work}
\label{app:related_work}

\paragraph{RLVR and trajectory sources.}
Beyond the methods discussed in the main text, Prefix-RFT supplies expert prefixes~\citep{huang2025prefixrft}, while ReLIFT, SRFT, and TRAPO combine RL with supervised updates~\citep{ma2025learning,fu2025srft,su2025trapo}. RePO and ExGRPO reuse successful historical trajectories~\citep{li2025repo,zhan2025exgrpo}, and RLEP draws experience from a stronger seed policy~\citep{zhang2025rlep}. These methods motivate enriching the learner's available signals. Our focus is the temporal choice of source: how much additional optimization should a teacher receive before guiding an earlier learner?

\paragraph{Self-distillation and temporal teacher selection.}
ReST and STaR bootstrap from successful generations~\citep{gulcehre2023reinforced,zelikman2022star}; context distillation uses privileged context~\citep{snell2022learning}. On-policy distillation evaluates teacher predictions on student-generated responses~\citep{agarwal2024policy,lu2025onpolicydistillation}, with OPSD and SDPO using reference solutions or feedback to inform those predictions~\citep{zhao2026opsd,hubotter2026reinforcement}. Self-distillation also supports continual learning~\citep{shenfeld2026selfdistillationenablescontinuallearning}. Temporal self-distillation instead obtains its teacher through further optimization along the learner's training path. NPO and NPD examine this teacher choice through trajectory-level and token-level transfer, respectively.

\section{AutoNPO Implementation and Pseudocode}
\label{app:autonpo_algorithm}

At step $t$, the current policy $\pi^{(t)}$ can serve as a future guide for an
earlier checkpoint $\pi^{(t-\Delta)}$. The controller first confirms that it
has acquired useful new capability, then selects a rollback distance using
$\hat{\mathcal{S}}(\Delta)=\hat Q(\Delta)/\hat V(\Delta)$.
The details below complement the overview in Section~\ref{sec:autonpo};
Algorithm~\ref{alg:autonpo} gives the complete procedure.

\paragraph{Mistake pool.}
AutoNPO maintains a lightweight mistake pool $\mathcal{B}$ containing prompts
with low recent group accuracy together with the steps at which they were
failed. The pool serves both as the probe set for detecting newly acquired
capability and as the evaluation set for comparing candidate rollback
distances. After an intervention is selected, the relevant subset also defines
the prompts on which future-self guidance is prepared.

\paragraph{Trigger.}
The controller uses a coarse-to-fine two-stage check. A zero-cost warning stage
monitors training reward and policy entropy, and activates when reward
stagnation coincides with continued entropy decline, a practical indicator of
reduced exploration. If this pattern persists for several probe intervals, a
confirmation stage rolls out the current policy once on a small subset of
$\mathcal{B}$. The resulting pass-rate determines whether the current policy
has acquired enough new capability to serve as a useful guide; the same rollout
also supplies the estimates of $\hat Q(\Delta)$ used below.

\paragraph{Rollback distance.}
Once an intervention is confirmed, AutoNPO selects
\begin{equation}
    \Delta^\star
    =
    \arg\max_{\Delta\in\mathcal{D}}
    \frac{\hat Q(\Delta)}{\hat V(\Delta)},
\end{equation}
where $\mathcal{D}$ is the set of available checkpoint distances.
For each candidate $\Delta$, let $\mathcal{B}_{\Delta}$ denote the subset of
mistake-pool prompts that failed during the segment beginning at
$t-\Delta$. We estimate
\begin{equation}
    \hat Q(\Delta)
    =
    \mathrm{pass\text{-}rate}
    \big(
        \pi^{(t)};\mathcal{B}_{\Delta}
    \big),
\end{equation}
and use an exponential proxy based on the per-token KL between
$\pi^{(t)}$ and $\pi^{(t-\Delta)}$ for $\hat V(\Delta)$.
Because all candidate $\hat Q(\Delta)$ values are obtained from the same
confirmation rollout, checkpoint ranking requires no additional inference
beyond the probe itself.

\paragraph{Execution.}
After choosing $\Delta^\star$, the current policy $\pi^{(t)}$ acts as the
future guide for the earlier checkpoint $\pi^{(t-\Delta^\star)}$. We roll out
$\pi^{(t)}$ on prompts in $\mathcal{B}_{\Delta^\star}$, verify the outputs,
and cache one correct trajectory per prompt when available. Training then
restores checkpoint $t-\Delta^\star$ and replays the intervened segment using
the NPO mechanism. During replay, cached guidance is injected only for prompts
in $\mathcal{B}_{\Delta^\star}$; all other prompts retain fully on-policy
rollout groups.

Once the replayed run catches back up to step $t$, the controller enters a
short cooldown and then re-arms. The resulting policy can therefore later
serve as a stronger future guide for another intervention, allowing temporal
self-transfer to repeat as training progresses. Algorithm~\ref{alg:autonpo}
gives the full procedure.

\begin{algorithm}[ht]
\caption{AutoNPO: adaptive intervention for near-future policy optimization}
\label{alg:autonpo}
\begin{algorithmic}[1]
\Require dataset $\mathcal{D}$, base policy $\pi^{(0)}$, mistake threshold $\tau_\mathrm{err}$, confirmation pass-rate threshold $\tau_\mathrm{pass}$, probe size $N_\mathrm{probe}$, probe interval $T_\mathrm{probe}$, cooldown length $T_\mathrm{cool}$, cost proxy $\hat V(\cdot)$
\State $\mathcal{B}\leftarrow\emptyset$, $\texttt{alert}\leftarrow 0$, $\texttt{in\_retro}\leftarrow\text{False}$, $\mathcal{G}\leftarrow\emptyset$, $t_\mathrm{cool}\leftarrow 0$
\For{$t=1,2,\ldots$}
  \State sample batch $\{x\}$ from $\mathcal{D}$
  \If{$\texttt{in\_retro}$} \textbf{for} each $x$ with $\mathcal{G}[x]$ defined, schedule $\mathcal{G}[x]$ to replace one rollout slot \EndIf
  \State roll out with $\pi^{(t)}$, compute rewards and group-relative advantages
  \State update $\pi^{(t+1)}$ via the clipped objective in Eq.~\eqref{eq:nfpo_obj}
  \State update reward EMA and entropy history
  \For{each prompt $x$ in the batch}
    \If{mean group accuracy of $x < \tau_\mathrm{err}$} record prompt $x$ in $\mathcal{B}$ with timestamp $t$ \EndIf
  \EndFor
  \If{not $\texttt{in\_retro}$ and $t \ge t_\mathrm{cool}$ and $t \bmod T_\mathrm{probe}=0$}
    \State \textcolor{gray}{\textit{// Stage 1: zero-cost warning}}
    \State $\texttt{alert}\leftarrow\texttt{alert}+1$ if reward stagnates and entropy declines, else $\texttt{alert}\leftarrow 0$
    \If{$\texttt{alert}\ge m$}
      \State \textcolor{gray}{\textit{// Stage 2: confirmation rollout}}
      \State sample $N_\mathrm{probe}$ prompts from $\mathcal{B}$; roll out with $\pi^{(t)}$
      \State $\hat p \leftarrow$ pass-rate on the probe sample
      \State \textbf{for} each saved checkpoint at distance $\Delta$: $\hat Q(\Delta) \leftarrow$ pass-rate of $\pi^{(t)}$ on $\mathcal{B}_\Delta$
      \If{$\hat p > \tau_\mathrm{pass}$}
        \State $\Delta^\star \leftarrow \arg\max_\Delta \hat Q(\Delta)/\hat V(\Delta)$ \Comment{Sec.~\ref{sec:autonpo}}
        \State $\mathcal{G}\leftarrow$ verified-correct re-rollout of $\pi^{(t)}$ on $\mathcal{B}_{\Delta^\star}$
        \State load model weights and dataloader state from checkpoint $t-\Delta^\star$
        \State $t_\mathrm{resume}\leftarrow t$; $\texttt{in\_retro}\leftarrow\text{True}$
      \EndIf
      \State $\texttt{alert}\leftarrow 0$
    \EndIf
  \EndIf
  \If{$\texttt{in\_retro}$ and current step $\ge t_\mathrm{resume}$}
    \State $\texttt{in\_retro}\leftarrow\text{False}$; $\mathcal{G}\leftarrow\emptyset$; $t_\mathrm{cool}\leftarrow t+T_\mathrm{cool}$
  \EndIf
\EndFor
\end{algorithmic}
\end{algorithm}

\section{Broader Impact}
\label{app:broader-impact}
Temporal self-distillation obtains guidance from the model's own later checkpoints, reducing dependence on stronger external teachers for post-training. Improved reasoning can support useful applications, but can also enable more effective misuse, including misleading content. Verifier-based training may reinforce harmful behavior if the verifier rewards it. Deployment therefore requires safety evaluation and safeguards appropriate to the application, alongside the reasoning improvements studied here.

\end{document}